\documentclass[10pt,twocolumn,letterpaper]{article}

\usepackage[pagenumbers]{cvpr} 

\usepackage[dvipsnames]{xcolor}

\definecolor{cvprblue}{rgb}{0.21,0.49,0.74}
\usepackage[pagebackref,breaklinks,colorlinks,citecolor=cvprblue]{hyperref}

\usepackage[accsupp]{axessibility} 

\def\paperID{106} 
\def\confName{CVPR}
\def\confYear{2024}

\title{Exploring AIGC Video Quality: A Focus on Visual Harmony, Video-Text Consistency and Domain Distribution Gap}

\author{Bowen Qu$^1$
\and
Xiaoyu Liang$^1$
\and
Shangkun Sun$^{1,2}$
\and
Wei Gao$^{1,2{\dagger}}$\\
\and
$^1$ School of Electronic and Computer Engineering, Peking University, China\\
$^2$ Peng Cheng Laboratory, China\\
{\tt\small \{bowenqu, 2000017789, sunshk\}@stu.pku.edu.cn, gaowei262@pku.edu.cn}
}

\begin{document}
\maketitle
\begin{abstract}
\renewcommand{\thefootnote}{}
\footnotetext{$^{\dagger}$Corresponding author.\ This work was supported by Natural Science Foundation of China (62271013, 62031013), Guangdong Province Pearl River Talent Program High-Caliber Personnel - Elite Youth Talent (2021QN020708), Shenzhen Science and Technology Program (JCYJ20230807120808017), and Sponsored by CAAI-MindSpore Open Fund, developed on OpenI Community (CAAIXSJLJJ-2023-MindSpore07).}

The recent advancements in Text-to-Video Artificial Intelligence Generated Content (AIGC) have been remarkable. Compared with traditional videos, the assessment of AIGC videos encounters various challenges: visual inconsistency that defy common sense, discrepancies between content and the textual prompt, and distribution gap between various generative models, etc.
Target at these challenges, in this work,
we categorize the assessment of AIGC video quality into three dimensions: visual harmony, video-text consistency, and domain distribution gap. For each dimension, we design specific modules to provide a comprehensive quality assessment of AIGC videos. Furthermore, our research identifies significant variations in visual quality, fluidity, and style among videos generated by different text-to-video models. Predicting the source generative model can make the AIGC video features more discriminative, which enhances the quality assessment performance. 
The proposed method was used in the \textbf{third-place} winner of the NTIRE 2024 Quality Assessment for AI-Generated Content - Track 2 Video, demonstrating its effectiveness. Code will be available at \href{https://github.com/Coobiw/TriVQA}{https://github.com/Coobiw/TriVQA}.
\end{abstract}
\section{Introduction}
\begin{figure}[!t]
\centering
\includegraphics[width=1\linewidth]{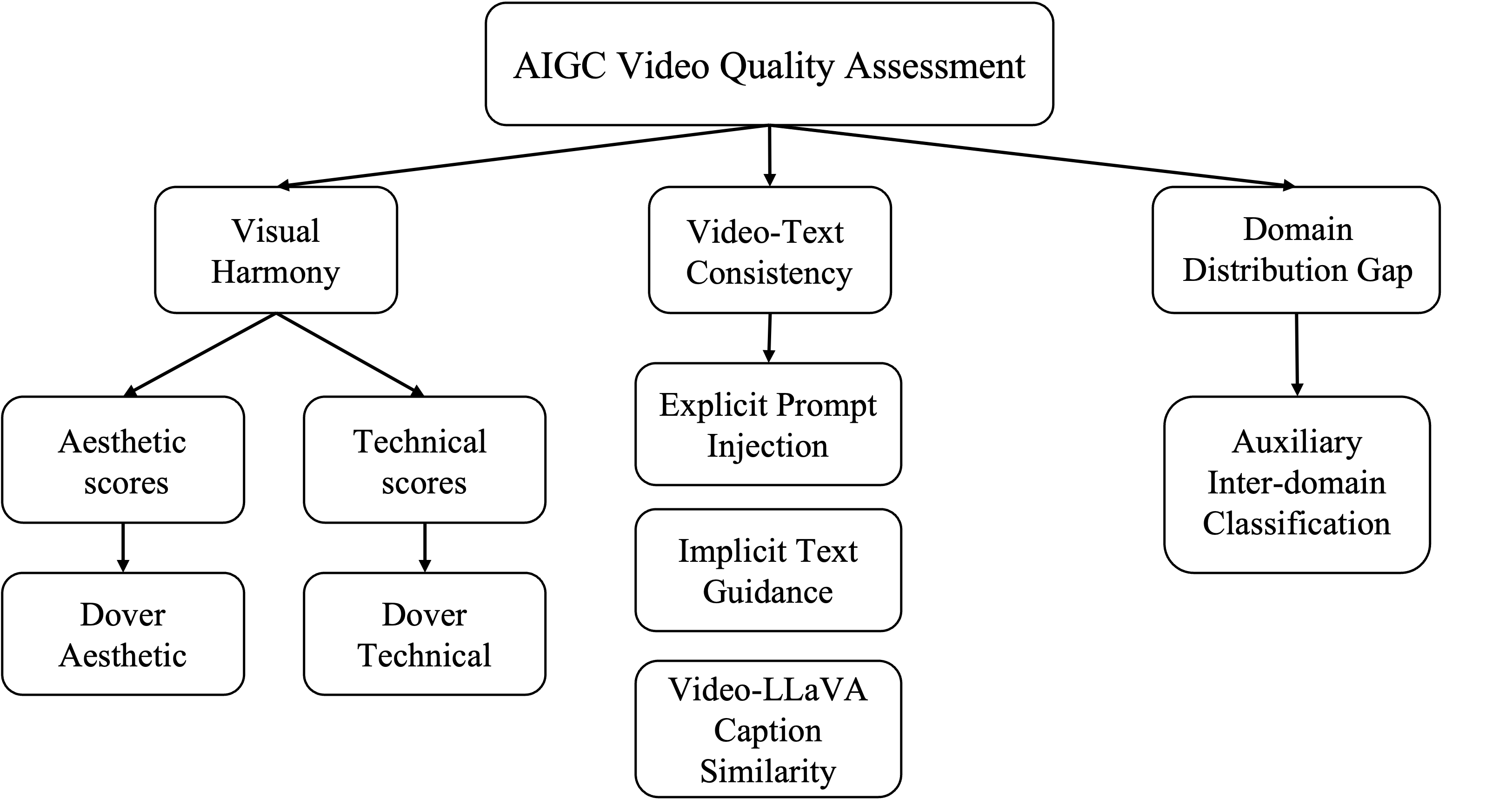}
\caption{Three Dimensions for AIGC Video Quality Assessment.}
\label{fig1}
\end{figure}

In recent years, the emergence of Artificial Intelligence Generated Content (AIGC), including images, texts and videos, has significantly influenced the digital media production landscape. Numerous video generation models based on different technical routes have been developed, which can be branched into GAN-based~\cite{Vondrick2016GeneratingVW}, autoregressive-based~\cite{yan2021videogpt} and diffusion-based~\cite{wang2023lavie}. As an emerging new video type, AI-generated video needs more comprehensive quality assessment (QA) for the users to get better visual experience. 

On the one hand, Inception Score (IS)~\cite{salimans2016improved}, Fréchet Video Distance (FVD)~\cite{unterthiner2019accurate} and Fréchet Inception Distance (FID)~\cite{hessel2021clipscore} are usually employed to evaluate the perceptual quality. On the other hand, CLIPScore~\cite{hessel2021clipscore} is usually used to evaluate the video-text correspondence. However, these metrics are heavily reliant on specific datasets or pretrained models, which is not comprehensive enough.

There are also many advanced  Video Quality Assessment (VQA) methods~\cite{wu2022fastvqa, kou2023stablevqa, zhao2023quality, sun2022a} proposed for evaluating natural or UGC (User Generated Content) videos. Nevertheless, due to the limited focus of these methods on issues like visual abnormalities in AIGC videos, their zero-shot effectiveness is not satisfactory. Additionally, there are significant differences in the content distributions generated by different models, posing a strong challenge to the robustness of the framework. Besides, most of them are single-modality based, focus on the technical or aesthetic visual quality. However, AIGC videos generated by text-to-video model are inherently multimodal entities, each accompanied by a corresponding textual prompt.
In reality, these models only take videos as input, which is not sufficient for a comprehensive evaluation due to the lack of understanding of entire textual prompts.
Some methods~\cite{wu2023bvqi, liu2023evalcrafter} take the text into consideration for more comprehensive assessment, leveraging the strong multi-modality ability of CLIP~\cite{radford2021learning}. They use hard prompts like "a \{high, low\} quality photo" instead, which is not a good way to evaluate the video-text correspondence.

In this work, we assess the AIGC videos quality from three dimensions: visual harmony, video-text consistency, and domain distribution gap. 
The overall framework is shown in Fig.~\ref{fig1}.
As for visual harmony, we refer to DOVER~\cite{wu2023dover} for the aesthetic and technical evaluation of the videos. To measure video-text consistency, we apply explicit prompt injection, implicit text guidance and caption similarity. We inject the corresponding prompts of the videos into the video features using Text2Video Cross Attention Pooling~\cite{qu2024ipiqa}. We also utilize BVQI's implicit text method~\cite{wu2023bvqi} and jointly optimize the evaluation network using both implicit text and explicit prompts. Building upon this, we utilize the video-text Multimodal Large Language Model (MLLM), Video-LLaVA~\cite{lin2023video} to generate additional captions for each video segment. We use Sentence-BERT~\cite{reimers-2019-sentence-bert} to get the embeddings of generated captions and given prompts, and then calculate cosine similarity between them to further optimize the network. To improve the spatio-temporal modeling capability, we also integrate strongly pretrained video backbones by linear-probing like UniformerV2~\cite{li2022uniformerv2} for model ensemble to get robust results.

Additionally, considering the domain distribution gaps in the videos generated by different text-to-video models, in supervised learning, we predict not only the final mean opinion score (MOS) but also which text-to-video model generated the video. This additional classification aids the model in better understanding video features. Experiments have shown that this significantly enhances the performance of our model.

To summarize, our contributions are three-fold:
\textbf{1)} We propose a new quality assessment framework for AIGC videos, which we  decouple into three aspects: visual harmony, video-text consistency and domain distribution gap. 
\textbf{2)} For each aspect, we design specific modeling methods such as LLM and auxiliary inter-domain classifiers, to propose effective solutions.
\textbf{3)} Our method shows remarkabale improvements on AIGC videos assessment and is used in the \textbf{third-place winner} of the NTIRE 2024 Quality Assessment for AI-Generated Content - Track 2 Video~\cite{ntire2024QA_AI, kou2024subjectivealigned}.

\section{Related work}
\label{sec:formatting}
\subsection{No-Reference Video Quality Assessment}
Quality Assessment has become a crucial task, and significant progress has been made across multiple domains~\cite{qu2024ipiqa, wu2023dover, 9428383, fan2023screenbased, 10422856, 10273743} in recent years. Classic NR-VQA methods adopt handcrafted features as evaluation metrics~\cite{BLIINDS,TLVQM,chipQA,tu2020ugc,Two-LevelQA,6353522}. These methods can extract useful information like color, motion and temporal-spatial features, while keeping low computational complexity. Some other methods~\cite{QAwild2019, wu2022fasterquality, wu2022fastvqa} extract video features by deep neural networks. DOVER~\cite{wu2023dover} categorizes video quality evaluation into aesthetic and technical dimensions, leveraging distinct backbones for feature extraction. Scores are assigned individually and then aggregated based on a predetermined ratio. By sampling novel fragments as input for deep neural networks and designing a fragment attention network base on Swin Transformers, Fast-VQA~\cite{wu2022fastvqa} efficiently retains quality-related information. StableVQA~\cite{kou2023stablevqa} measures video stability by obtaining optical flow, semantic, and blur features separately. Recently, some novel methods use visual-language pre-training models to evaluate videos. Q-Align~\cite{wu2023qalign} utilizes the comprehension abilities of a Multimodal Large Language Model to transform the video quality evaluation task into the generation of discrete quality level words. BVQI~\cite{wu2023bvqi} introduces the text-language model CLIP to evaluate video quality by assessing the affinity between positive or negative prompts and extracted frames.

\subsection{Video Generation and Quality Assessment}
Video generation aims to achieve videos with high visual quality and consistent, smooth movements that closely approximate the real world. Image generators based on Generative Adversarial Networks (GANs)~\cite{Goodfellow2022GAN} have been extended to be effectively used for video generation. However, these methods~\cite{Vondrick2016GeneratingVW, Saito2016TemporalGA, Wang2019G3ANDA} often encounter issues with mode collapse, leading to lower quality and stability in the generated content. Additionally, some approaches~\cite{rakhimov2020latent, ge2022long, weissenborn2020scaling, yan2021videogpt} have proposed the use of auto-regressive models to learn the distribution of video data. These methods are capable of generating high-quality and stable videos, but they require a significantly high computational cost. Recent methods~\cite{ho2020denoising, harvey2022flexible, ho2022video, mei2022vidm, blattmann2023align, wang2023lavie, chen2023seine} in video generation have predominantly focused on diffusion models, achieving very promising results.

Several quantitative evaluation metrics have been proposed for AI-generated videos, mainly focusing on assessing perceptual quality and the video-text correspondence. For perceptual quality, Inception Score (IS)~\cite{salimans2016improved}, Fréchet Video Distance (FVD)~\cite{unterthiner2019accurate} and Fréchet Inception Distance (FID)~\cite{hessel2021clipscore} are usually employed. CLIPScore~\cite{hessel2021clipscore} is mainly used to evaluate the video-text correspondence, leveraging the capabilities of CLIP~\cite{radford2021learning}. However, these methods are heavily reliant on specific datasets or pretrained models and sensitive to their calculation parameters, such as batch size. Additionally, they do not take the human visual system into consideration, which mean mis-alignment with human perception in assessing AI-generated video.
\section{Method}
\begin{figure*}[!t]
\centering
\includegraphics[width=1\linewidth]{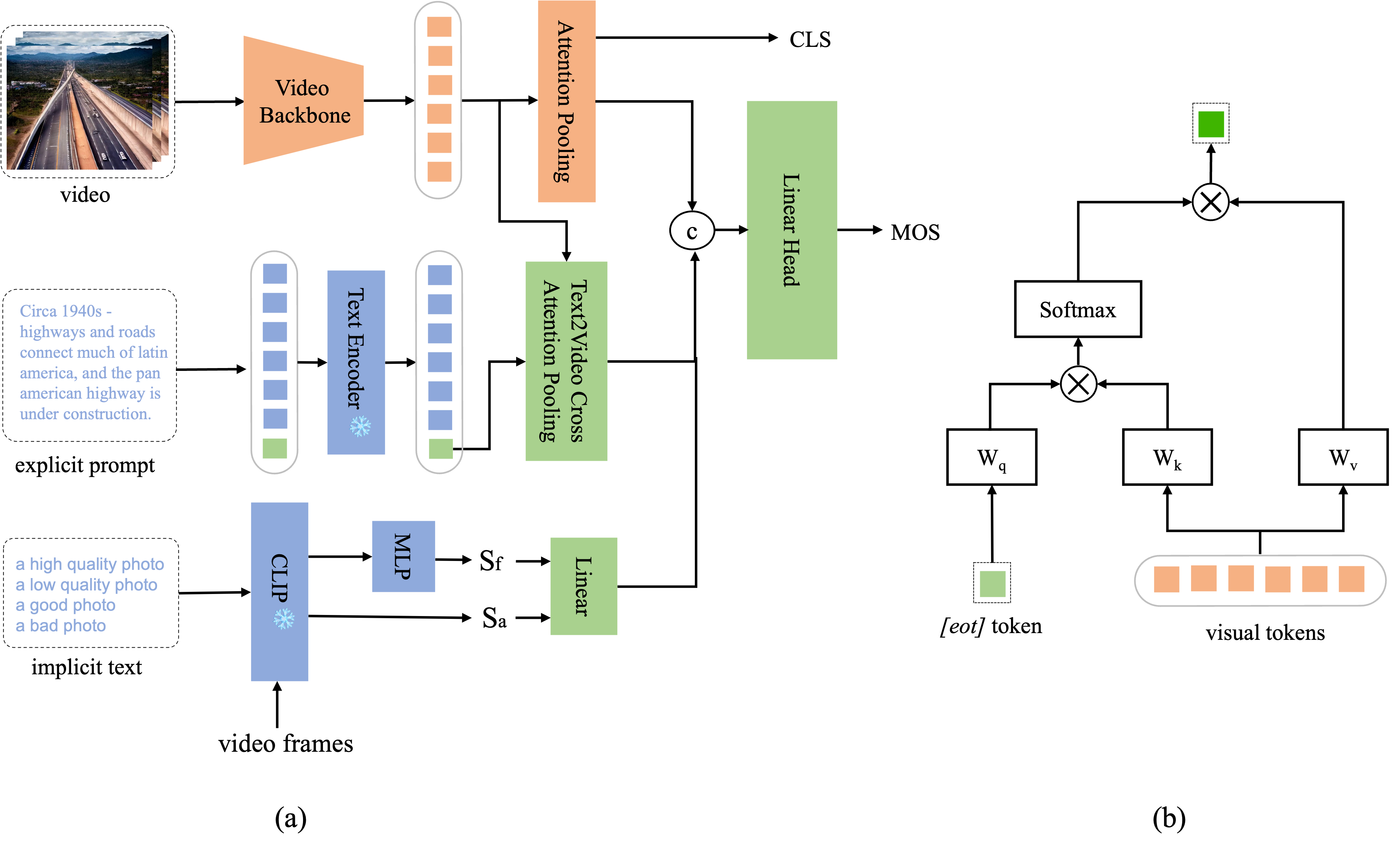}
\caption{Detailed overview of our framework. (a) illustrates the whole framework, which serves as an incremental enhancement for DOVER. Except for the visual part, our framework also incorporates modules to deal with the explicit prompt and implicit text, enriching the capability in video-text consistency assessment. (b) shows the workflow of the Text2Video Cross Attention Pooling module, which is based on cross-attention mechanism.}
\label{fig2}
\end{figure*}

The overview of our proposed method is shown in Fig. \ref{fig2}. 
Our approach proposes solutions from various aspects including Visual Harmony, Video-Text Consistency, Domain Distribution Gap, etc., which we will elaborate on in the following sections.
In summary, it serves as a dual-stream architecture to simultaneously process the AIGC videos and corresponding textual prompts. We think that the inductive bias of this framework matches the multi-modal nature of AIGC videos better. In order to cooperate with the visual harmony model, modules for explicit prompt and implicit text processing are injected, serving as an incremental enhancement for existing VQA methods. Specifically speaking, the video backbone can be initialized by DOVER~\cite{wu2023dover}.

\subsection{Visual Harmony}
Due to the impressive performance of DOVER~\cite{wu2023dover} across multiple Video Quality Assessment datasets~\cite{konvid1k, hosu2017konstanz, Sinno_2019, Ying_2021}, we select it for visual harmony modeling. DOVER consists of two branches: the aesthetic branch and the technical branch. These utilize the ConvNext~\cite{liu2022convnet} and Swin-Transformer~\cite{liu2021swin} backbone respectively, trained on their DIVIDE-3k dataset~\cite{wu2023dover}. The processing of input videos by these two branches differs. Notably, the technical branch exhibits additional patchifying and fragment sample operations, focusing more on patch-wise features and temporal information. We replace the non-learnable Global Average Pooling (GAP) layer with a learnable Attention Pooling layer (as shown in Fig. \ref{fig2}. Specifically, the output of the GAP is treated as the query, while the spatio-temporally flattened visual tokens are considered as keys and values for cross-attention operations, collectively forming the Attention Pooling module.

\subsection{Video-Text Consistency}
AIGC videos possess multi-modality nature inherently because of their corresponding textual prompts which is the condition for text-to-video generative model. Due to this characteristic, we propose a multi-modal framework, integrated with explicit textual prompt and implicit text with hard template. These operations enable our model to take textual prompts into consideration and acquire more comprehensive multi-modal features, following text-video interaction. In order to incorporate video-text consistency capabilities into our video quality assessment framework more directly, we also employ the strong and robust video-text Multimodal Large Language Model (MLLM), Video-LLaVA, to generate captions by the input videos. Then we calculate sentence embedding similarity with the respective textual prompts for a direct zero-shot video-text consistency score.

\paragraph{Explicit Prompt Injection.}
AIGC videos have inherent multimodal natures from birth. The explicit prompt is a specific condition and guidance for the text-to-video model to generate the corresponding video. Meanwhile, video-text consistency is also an important degree of AIGC video quality assessment. So, we produce a CLIP-like dual-stream architecture with two separate encoders to process the video and prompt respectively. Given a video $V$ and the corresponding explicit prompt $T$, let $f_{\theta_{\text{v}}}(V)$ represent the video embedding, extracted by the video encoder with parameters $\theta_{\text{v}}$, and let $h_{\theta_{\text{t}}}(T)$ represent the text embedding produced by the text encoder with parameters $\theta_{\text{t}}$. The embedding of \textit{[eot]} (end of text) token to represent the entire prompt. The process is shown as following, where $F_v$ and $F_t$ refer to the video and text features respectively.$F_{eot}$ refers to the \textit{[eot]} token embedding, which serves as the global encoding of the whole textual prompt.
\begin{equation}
\begin{aligned}
F_v &= f_{\theta^{\prime}_{\text{v}}}(I), \\
F_t &= h_{\theta_{\text{t}}}(T, \textit{[eot]}), \\
F_{eot} &= F_t[:,-1,:].
\end{aligned}
\end{equation}

After separate feature extractions, explicit prompt embedding needs to interact with the visual embeddings. We design Text2Video Cross Attention Pooling for this, which is based on cross-attention mechanism. The \textit{[eot]}, respresentation of the emplicit prompt, serves as the query. As for the visual embeddings, the output of the video backbone, we flatten them on the spatio-temporal dimensions and serve them as the key and value.

\begin{equation}
\begin{aligned}
Q &= W_q \cdot F_{eot},  \\
K &= W_k \cdot F_v, \\
V &= W_v \cdot F_v,
\end{aligned}    
\end{equation}

\noindent where $W_q$, $W_k$, $W_v$ refer to the query, key, value projection matrix and $Q$, $K$, $V$ refer to the query, key, value respectively. Then, scaled-dot attention~\cite{vaswani2023attention} is calculated.

\begin{equation}
\text{Attention}(Q, K, V) = \text{softmax}\left(\frac{QK^T}{\sqrt{d_k}}\right)V,
\end{equation}

\noindent where $d_k$ is the number of hidden state channels. The workflow of Text2Video Cross Attention Pooling module is shown in Fig. \ref{fig2} (b). After these procedures, we gain the text-video embedding, which can serve as a significant supplement for AIGC video quality assessment pipeline and work well on the video-text consistency evaluation.

\paragraph{Implicit Text Guidance.}
Inspired by BVQI ~\cite{wu2023bvqi}, we use an implicit text module to evaluate video quality. We calculate the affinity scores between a given $N$ frames video ($V$) and two pairs of texts ($T_{0}, T_{1}$), where each pair consists of one positive text and one negative text, along with the feature score of the video. Subsequently, these scores are aggregated through a linear output.

The sampling pipeline is different from the one proposed in ~\cite{wu2023bvqi}. In order to fully explore the potential information of the video, all frames are utilized and cropped to the size of 224 $\times$ 224, ensuring the integrity and efficient utilization of the information. Then, following ~\cite{wu2023bvqi}, we use CLIP~\cite{pmlr-v139-radford21a} visual ($E_{v}$) and textual ($E_{t}$) encoders to extract the video feature of frame $i$ ($f_{v,i}$) and the text feature of text pair $j$ ($f_{t,j}$), calculate the affinity and conduct sigmoid remapping to form the final affinity scores $S_{a,0}$ and $S_{a,1}$:
\begin{equation}
  f_{v,i}=E_{v}(V_{i}),
\label{eq:important}
\end{equation}
\begin{equation}
  f_{t,j}={E_{t}(T_{j,pos}),E_{t}(T_{j,neg})},
  \label{eq:important}
\end{equation}
\begin{equation}
S_{a,j}=Sigmoid(\frac{\sum_{i=0}^{N-1}(f_{v,i}\cdot f_{t,j,pos}^T-f_{v,i}\cdot f_{t,j,neg}^T)}{N}).
  \label{eq:important}
\end{equation}

To bridge the gap between AIGC video frames and reality images, we generate feature score $S_{f}$ from the generated video features and project it within the range of [0,1]:
\begin{equation}
  S_{f}=Sigmoid(GELU(MLPs(f_{v}))).
  \label{eq:important}
\end{equation}

Ultimately, we combine it with affinity scores to produce implicit text score through a linear output.
\begin{equation}
  Score=Linear(S_{f},S_{a,0},S_{a,1}).
  \label{eq:important}
\end{equation}

\paragraph{Video-LLaVA Caption Similarity.}
The modeling of video-text consistency above is implicit. We also want to insert it into our assessment explicitly. So we use a video captioning model to translate the AIGC video into a brief description. Considering that MLLM has strong capabilities especially on zero-shot and few-shot learning, we choose Video-LLaVA~\cite{lin2023video} to generate appropriate caption for the corresponding AIGC video. 

\begin{figure}[!h]
\centering
\includegraphics[width=1\linewidth]{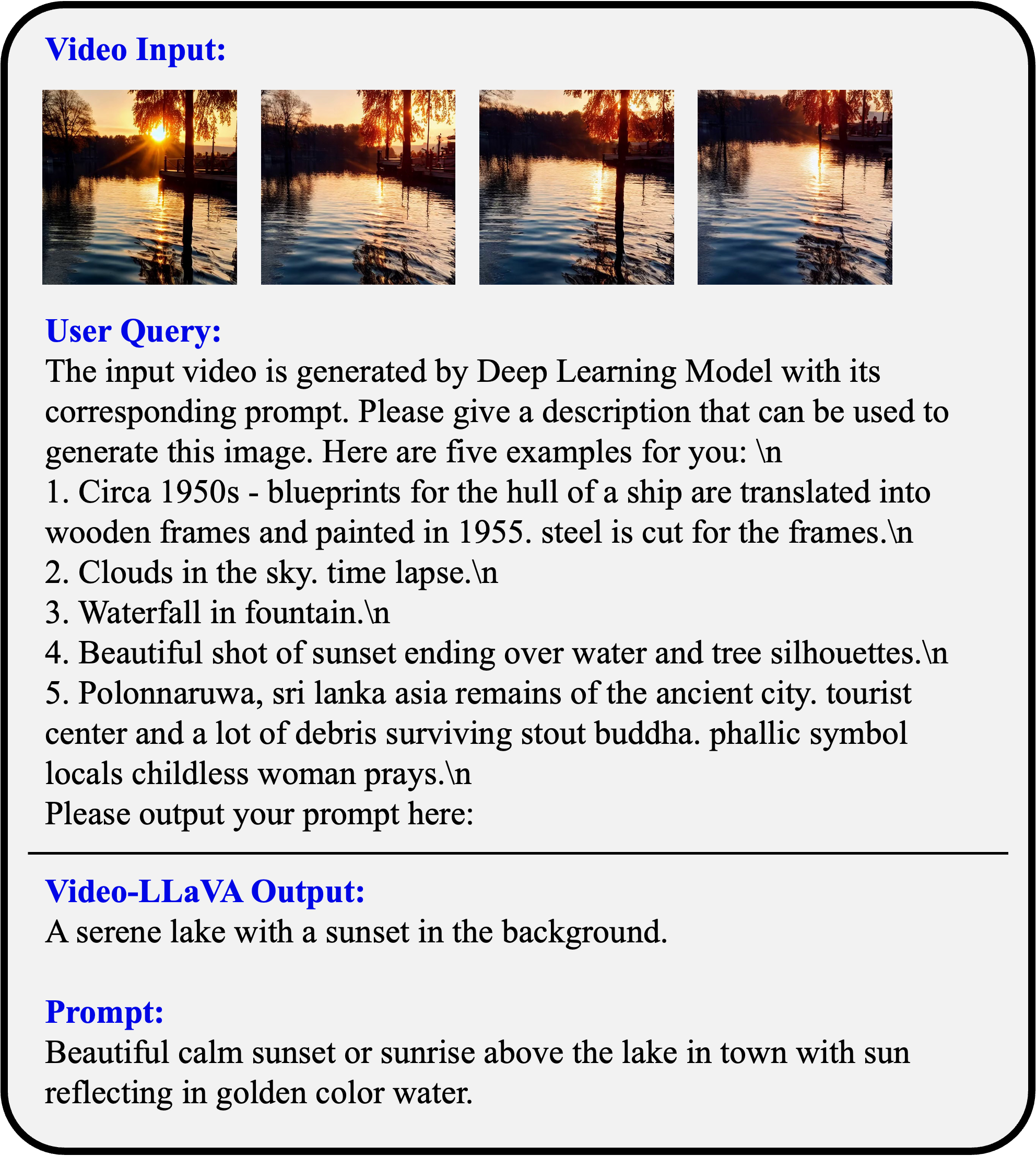}
\caption{One example used in In-Context-Learning for Video-LLaVA to generate the prompt-like caption.}
\label{fig3}
\end{figure}

However, the style of natural captions is different from the textual prompts of text-to-video model. In order to alleviate this problem, we want to leverage the in-context learning ability of Video-LLaVA. So, we use 5-shot inference. In practice, we randomly choose 5 textual prompts from the train dataset, which serve as the context of MLLM. Fig. \ref{fig3} shows the workflow of Video-LLaVA inference.

After that, we use Sentence-BERT~\cite{reimers-2019-sentence-bert} to extract the embeddings of generated captions and corresponding textual prompts and calculate cosine similarity. We normalize the output and serve them as the finale caption similarity scores.

\subsection{Auxiliary Inter-domain Classification}
Video generation models typically follow three technical approaches: GAN-based, auto-regressive based, and diffusion-based methods. Videos produced by these varying models exhibit distinctions in visual quality, fluency and style. Predicting the specific generative model behind AIGC videos can lead to the extraction of more discriminative features. This capability significantly aids in the enhanced assessment of AIGC video quality. Consequently, we have integrated an additional auxiliary inter-domain classification branch. This component predicts the origin of given AIGC videos from among 10 potential video generation models, substantially benefiting the quality evaluation of AIGC videos. We use cross-entropy loss $L_{cls}$ as the auxiliary objective function, incorporating it into the main loss with a weight of $\beta$.

\begin{align}
    L &= L_{qual} + \beta \cdot L_{cls} \nonumber \\
      &= L_{plcc} + \alpha \cdot L_{rank} + \beta \cdot L_{cls},
\end{align}

\noindent where $L_{qual}$ refers to the quality loss, composed of PLCC(Pearson Linear Correlation Coefficient) loss $L_{plcc}$ and rank loss~\cite{gao2019learning} $L_{rank}$ with $\alpha$ as its weight. In practice, the values of $\alpha$ and $\beta$ are set to 0.3 and 0.2 respectively.

\section{Experiments}
\subsection{Dataset}
Our experiments utilize the AI-Generated video dataset proposed for the video track of the NTIRE 2024 Quality Assessment for AI-Generated Content~\cite{ntire2024QA_AI, kou2024subjectivealigned}. The dataset can be divided into three parts: the validation dataset, the test dataset, and the train dataset. The train dataset comprises 7000 videos, along with their corresponding Mean Opinion Scores (MOS) and textual prompts. The validation and test datasets encompass 2000 and 1000 videos respectively, including only their prompts.

All videos comprise either 15 or 16 frames, possessing a duration of 4 seconds, and exhibit a frame rate of either 3.75 or 4 frames per second. We observe that video filenames in the train dataset adhere to a structured format: "x\_y.mp4," where "x" identifies the video's unique number within the dataset, and "y" indicates the distinct generative model employed.

\subsection{Implement Details}
Following DOVER~\cite{wu2023dover}, we apply temporal and spatial sample to raw videos. During both training and testing phases, frames are sampled comprehensively across all branches. Specifically, in the DOVER technical branch, 7 $\times$ 7 spatial grids are utilized. For other branches, frames are sampled and resized to a resolution of 224 $\times$ 224.

The whole procedures are implemented using Python programming language, leveraging the PyTorch framework for deep learning. For dover-base branches, we employ the AdamW optimizer with a weight decay of 0.05. Different learning rates are used for the backbones and heads: the backbones are trained with a learning rate of 6.25e-5, while the heads are trained with a larger learning rate of 6.25e-4. The training process is conducted on a single NVIDIA GeForce RTX 4090 24GB GPU, with each branch trained for 25 epochs. We divide the 25 training epochs into two phases: 10 linear-probe epochs and 15 end-to-end fine-tuning epochs, following the DOVER~\cite{wu2023dover} approach. Applying this training strategy to the branches results in a training time of approximately 8 hours. Due to the limitation of GPU memory, branches with larger backbones requires 25 linear-probe epochs, which takes approximately 4 hours to complete.

In order for further improvement, we also use model-ensemble tricks, by weighted summation of results from various models. The advanced VQA methods~\cite{zhao2023quality, wu2022fastvqa, wu2022fasterquality} and linear-probing of strongly pretrained backbones~\cite{li2022uniformerv2, wang2022masked, li2023unmasked} are integrated.

\subsection{Evaluation Metrics}
Like traditional No-Reference Video Quality Assessment, MOS (Mean Opinion Score) is the ground-truth. The mixture of PLCC (Pearson’s Linear Correlation Coefficient) and SROCC (Spearman’s Rank Order Correlation Coefficient) serves as the evaluation metrics. 

PLCC is a measure of the linear correlation between predicted scores and MOS. It ranges from -1 to 1, where 1 is total positive linear correlation, 0 is no linear correlation, and -1 is total negative linear correlation. The formula is shown as following:

\begin{equation}
PLCC = \frac{\sum (X_i - \overline{X})(Y_i - \overline{Y})}{\sqrt{\sum (X_i - \overline{X})^2 \sum (Y_i - \overline{Y})^2}},
\end{equation}

\noindent where $X_i$ and $Y_i$ refer to the prediction and target of the $i^{th}$ sample. $\overline{X}$ and $\overline{Y}$ are the means.

SROCC is a measure used to evaluate the strength and direction of association between two ranked variables. Unlike PLCC, which assesses linear relationships, SROCC is used to identify monotonic relationships (whether linear or not). It ranges from -1 to 1, where 1 indicates a perfect positive association, -1 a perfect negative association, and 0 no association.

\begin{equation}
    SROCC = 1 - \frac{6 \sum d_i^2}{n(n^2 - 1)},
\end{equation}

\noindent $d_i$ is the difference between the ranks of corresponding values of predictions and targets, and $n$ is the number of observations.

The overall score, \textit{MainScore}, is obtained by ignoring the sign and reporting the average of absolute values $((PLCC + SROCC)/2)$.

\subsection{Experimental Results}
We conduct a comparative evaluation of our method against fine-tuned state-of-the-art VQA methods including SimpleVQA~\cite{sun2022a}, BVQA~\cite{li2022blindly}, Fast-VQA~\cite{wu2022fastvqa} and DOVER~\cite{wu2023dover} on the validation dataset. To ensure a fair comparison, we report only the performance of a single model and do not employ model ensemble. The experimental results shown in Tab.\ref{tab1} demonstrate that our approach outperforms the existing VQA methods.

\begin{table}[!t]
\centering
\caption{Results on the NTIRE 2024 Quality Assessment for AI-Generated Content - Track 2 Video Challenge Validation.}
\label{tab1}
\begin{tabular}{c|ccc}
\hline
Models                     & PLCC & SROCC & Main Score \\
\hline
SimpleVQA~\cite{sun2022a} & 0.6338 & 0.6275 & 0.6306 \\
BVQA~\cite{li2022blindly} & 0.7486 & 0.7390 & 0.7438 \\
Fast-VQA~\cite{wu2022fastvqa} & 0.7295 & 0.7173 & 0.7234 \\
DOVER~\cite{wu2023dover} & 0.7693 & 0.7609 & 0.7651 \\
\hline
Ours & \textbf{0.8099}   & \textbf{0.7905}    & \textbf{0.8002} \\
\hline
\end{tabular}
\end{table}

The NTIRE 2024 Quality Assessment for AI-Generated Content - Track 2 Video Challenge has the goal of developing a solution for AIGC video quality assessment. 13 teams were involved in the finale submission stage. All these teams have evaluated their proposed methods on the unseen validation and test set, and then submit the fact-sheet. Tab. \ref{tab2} shows the leaderboard of this challenge, according to the \textit{MainScore} on test set. Our proposed method was used in the third-place winner of the NTIRE 2024 Quality Assessment for AI-Generated Content - Track 2 Video.~\cite{ntire2024QA_AI, kou2024subjectivealigned}

\begin{table}[!h]
\centering
\caption{The leaderboard of the NTIRE 2024 Quality Assessment for AI-Generated Content - Track 2 Video Challenge.}
\label{tab2}
\begin{tabular}{c|c}
\hline
Team name                     & Main Score \\
\hline
ICML-USTC                     & 0.8385    \\
Kwai-kaa                      & 0.824     \\
\textbf{SQL}                           & \textbf{0.8232}    \\
musicbeer                     & 0.8231    \\
finnbingo                     & 0.8211    \\
PromptSync                    & 0.8178    \\
QA-FTE                        & 0.8128    \\
MediaSecurity\_SYSU\&Alibaba & 0.8124    \\
IPPL-VQA                      & 0.8003    \\
IVP-Lab                       & 0.7944    \\
Oblivion                      & 0.7869    \\
CUC-IMC                       & 0.7802    \\
UBC DSL Team                  & 0.7531    \\
\hline
\end{tabular}
\end{table}

\subsection{Ablation Study}
\begin{table*}[!t]
\centering
\caption{The ablation results on the validation set.}
\label{tab3}
\resizebox{1\linewidth}{!}{
    \setlength\tabcolsep{3pt}
    \tiny
\begin{tabular}{@{}c c c c c|ccc@{}}
\toprule
Explicit-Prompt & Implicit-Text & Aux-Cls  &  Model-Ensemble & Video-LLaVA & PLCC     & SROCC     & \textbf{MainScore} \\ \midrule
                &               &            &                    &             & 0.7649   & 0.7417    & 0.7533             \\ \midrule
    \checkmark  &               &            &                    &             & 0.7888   & 0.7676    & 0.7782             \\ \midrule
                &   \checkmark  &            &                    &             & 0.7843   & 0.7631    & 0.7737             \\ \midrule
    \checkmark  &   \checkmark  &            &                    &             & 0.7991   & 0.7803    & 0.7897             \\ \midrule
    \checkmark  &               & \checkmark &                    &             & 0.8020   & 0.7814    & 0.7917             \\ \midrule
    \checkmark  &   \checkmark  & \checkmark &                    &             & 0.8099   & 0.7905    & 0.8002             \\ \midrule
    \checkmark  &   \checkmark  & \checkmark &   \checkmark       &             & 0.8317   & 0.8153    & 0.8235             \\ \midrule
    \checkmark  &   \checkmark  & \checkmark &   \checkmark       &  \checkmark & \textbf{0.8341}   & \textbf{0.8165}    & \textbf{0.8253}             \\ \bottomrule
\end{tabular}}
\end{table*}

To verify the effectiveness of proposed methods, we conduct ablation study on the validation set. We use the visual harmony model, i.e. fine-tuned DOVER~\cite{wu2023dover} as our baseline. The purpose of the ablation studies is to explore the effectiveness of explicit prompt injection, implicit text guidance, auxiliary inter-domain classification (represented by Aux-Cls), model ensemble and Video-LLaVA~\cite{lin2023video} caption similarity. Main results are shown in Tab. \ref{tab3}.

\noindent\textbf{Impact of \textit{Explicit Prompt Injection}}. The \textit{Explicit Prompt Injection} is designed to get multi-modal text-video interaction features for better assessment on video-text consistency. With \textit{Explicit Prompt Injection}, the performance increases from 0.7533 to 0.7782. This result shows that video-text consistency is important for AIGC video quality assessment and our \textit{Explicit Prompt Injection} can enhance the multi-modal understanding.

\noindent\textbf{Impact of \textit{Implicit Text Guidance}}. The \textit{Implicit Text Guidance} module leverges the multi-modal alignment and understanding capabilities of CLIP~\cite{radford2021learning} to enhance the features of video frames. With \textit{Implicit Text Guidance}, the performance increases 0.0204, compard with the visual harmony baseline.

\noindent\textbf{Impact of \textit{Auxiliary Inter-domain Classification}}. \textit{Auxiliary Inter-domain Classification} is an auxiliary task to predict the video generation model. AIGC videos generated by different text-to-video generative models have different visual quality, fluency and style. So, It is a good way to make the AIGC video features more discriminative. According to Tab. \ref{tab3} Line.2 and Line.5, using \textit{Auxiliary Inter-domain Classification} improves the \textit{MainScore} from 0.7782 to 0.7917. Additionally, comparing Tab. \ref{tab3} Line.4 with Line.6, there is an increase from 0.7897 to 0.8002. These results shows that our \textit{Auxiliary Inter-domain Classification} task benefits for AIGC video quality assessment.

\noindent\textbf{Impact of \textit{Model Ensemble}}. Model ensemble is a good way to make our results more robust. In order for further enhancement, we leverage the train pipeline shown in Fig. \ref{fig2}, initializing the video backbone by UniformerV2~\cite{li2022uniformerv2} etc. and linear-probing without auxiliary inter-domain classification loss. These video backbones are pretrained on huge database with a large number of videos. So, they have strong capability on spatio-temporal modeling. As shown in Tab. \ref{tab3}, \textit{Model Ensemble} can increase the performance from 0.8002 to 0.8235.

\noindent\textbf{Impact of \textit{Video-LLaVA Caption Similarity}}. In order to integrate the video-text consistency modeling explicitly, we use Video-LLaVA~\cite{lin2023video} to generate captions and calculate the cosine similairy between generated captions and textual prompts via Sentence-BERT~\cite{reimers-2019-sentence-bert}. We also give five prompts from the train dataset as context for Video-LLaVA to generate prompt-like caption, which is shown in Fig. \ref{fig3}. By this operation, \textit{MainScore} is increased from 0.8235 to 0.8253. We believe that the strong capabilities on in-context learning and zero-shot inference of MLLM will help a lot on the test dataset and other open scenes.

\section{Conclusion}
We decouple AIGC videos quality assessment into three dimensions: visual harmony, video-text consistency and domain distribution gap. According to this, we design corresponding models or modules respectively for comprehensive AIGC videos quality assessment. Due to the inherent multi-modal nature of AIGC videos, we propose a multi-modal framework, integrated with explicit and implicit textual prompts. During this research, we also find that videos generated by different text-to-video models have different visual quality, style and temporal fluency. Therefore, we incorporate an auxiliary inter-domain classification, predicting the source video generation model. This operation makes the features of AIGC videos more discriminative and benefits the quality assessment. Our method was used in the \textbf{third-place winner} of the  NTIRE 2024 Quality Assessment for AI-Generated Content - Track 2 Video. Experimental results show the effectiveness of our proposed method. We believe that AIGC videos quality assessment can give a beneficial feedback to text-to-video generation and a larger AIGC videos dataset with samples from more recent T2V models is needed.
{
    \small
    \bibliographystyle{ieeenat_fullname}
    \bibliography{main}
}


\end{document}